\documentclass[conference]{IEEEtran}
\IEEEoverridecommandlockouts
\usepackage{cite}
\usepackage{amsmath,amssymb,amsfonts}
\usepackage{graphicx}
\usepackage{textcomp}
\usepackage{xcolor}
\usepackage{algorithm}
\usepackage{algpseudocode}
\usepackage[numbers]{natbib}
\usepackage{amsmath}
\usepackage{amsfonts}
\usepackage{mathrsfs}
\usepackage{bbm}

\def\bbR{{\mathbb{R}}}
\def\bbS{{\mathbb{S}}}

\def\calO{{\mathcal{O}}}

\def\A{{\mathbf{A}}}

\def\G{{\mathbf{G}}}
\def\H{{\mathbf{H}}}
\def\I{{\mathbf{I}}}

\def\Q{{\mathbf{Q}}}

\def\a{{\mathbf{a}}}

\def\e{{\mathbf{e}}}

\def\g{{\mathbf{g}}}

\def\u{{\mathbf{u}}}

\def\w{{\mathbf{w}}}
\def\x{{\mathbf{x}}}
\def\y{{\mathbf{y}}}
\def\z{{\mathbf{z}}}

\def\defeq{\overset{\rm{def}}{=}}

\usepackage{amsthm}
\usepackage{etoolbox}
\theoremstyle{plain}
\newtheorem{theorem}{Theorem}

\newtheorem{lemma}{Lemma}
\newtheorem{assumption}{Assumption}

\DeclareMathOperator*{\argmax}{arg\,max}

\newenvironment{proofsketch}{\noindent\textit{Proof Sketch.}}{\qed}
\def\BibTeX{{\rm B\kern-.05em{\sc i\kern-.025em b}\kern-.08em
    T\kern-.1667em\lower.7ex\hbox{E}\kern-.125emX}}
\begin{document}

\title{Multiple Greedy Quasi-Newton Methods for Saddle Point Problems} 
\author{\IEEEauthorblockN{1\textsuperscript{st} Minheng Xiao}
\IEEEauthorblockA{\textit{dept. Integrated System Engineering} \\
\textit{Ohio State University}\\
Columbus, USA \\
minhengxiao@gmail.com}
\and
\IEEEauthorblockN{2\textsuperscript{nd} Zhizhong Wu}
\IEEEauthorblockA{\textit{dept. Engineering} \\
\textit{UC Berkeley}\\
Berkeley, USA \\
ecthelion.w@gmail.com}}

\maketitle

\begin{abstract}
This paper introduces the Multiple Greedy Quasi-Newton (MGSR1-SP) method, a novel approach designed to solve strongly-convex-strongly-concave (SCSC) saddle point problems. Our method enhances the approximation of the square of the indefinite Hessian matrix inherent in these problems, significantly improving both the accuracy and efficiency through iterative greedy updates. We provide a thorough theoretical analysis of MGSR1-SP, demonstrating its linear-quadratic convergence properties. Numerical experiments conducted on AUC maximization and adversarial debiasing problems, compared with state-of-the-art algorithms, underscore our method's enhanced convergence rates and superior quality in inverse Hessian estimation. These results affirm the potential of MGSR1-SP to improve performance across a broad spectrum of machine learning applications where efficient and accurate Hessian approximations are crucial.
\end{abstract}


\section{Introduction}
The saddle point problem is a fundamental formulation in machine learning and optimization and naturally emerges in several applications including game theory~\citep{du1995minimax, ricceri2013minimax}, robust optimization~\citep{leyffer2020survey, gabrel2014recent}, reinforcement learning~\citep{azar2017minimax, agarwal2020model}, AUC maximization~\citep{yang2023minimax, liu2021quasi, liu2022quasi}, fairness-aware machine learning~\citep{liu2021quasi, jin2023fairness}, and generative adversarial networks (GANs)~\citep{goodfellow2020generative}. In this paper, we consider the following saddle point problem formulated as
\begin{align}
\label{eq: objective function}
\min_{\x \in \mathbb{R}^{n_\x}}\max_{\y \in \mathbb{R}^{n_\y}} f(\x, \y),
\end{align}
where $f(\x, \y)$ is smooth, strongly-convex in $\x$, and strongly-concave in $\y$. The objective is to find the saddle point $(\x^*, \y^*)$ such that:
\begin{align*}
f(\x^*, \y) \leq f(\x^*, \y^*) \leq f(\x, \y^*)
\end{align*}
for all $\x \in \bbR^{n_\x}, \y \in \bbR^{n_\y}$.

Several first-order optimization techniques have been developed to solve saddle point problems with linear convergence rate of iteration complexity of $\calO(1/\epsilon)$ including the extragradient (EG) method~\citep{korpelevichextragradient, tseng1995linear}, the optimistic gradient descent ascent (OGDA) method~\citep{daskalakis2017training, popov1980modification}, the proximal point method~\citep{rockafellar1976monotone}, mirror-prox method~\citep{juditsky2011solving}, and dual extrapolation method~\citep{nesterov2006solving}. Their stochastic extensions have also been studied in large-scale 
settings~\citep{alacaoglu2022stochastic, chavdarova2019reducing, luo2019stochastic, luo2021near, palaniappan2016stochastic, tominin2021accelerated}.

Second-order methods enhance convergence in saddle point problems but often entail higher computational demands. The cubic regularized Newton (CRN) method, which achieves quadratic local convergence, requires the computation of the exact Hessian matrix and solving of cubic variational inequality sub-problems~\citep{huang2022cubic}. Other adaptations, like the Newton proximal extragradient~\citep{gonccalves2020iteration, sicre2020complexity}, the mirror-prox algorithm~\citep{bullins2022higher}, and the second-order optimistic method~\citep{jiang2022generalized}, incorporate line searches for step size optimization. Conversely, methods such as \citep{adil2022optimal} and \citep{lin2024perseus} avoid such complexities by omitting line searches and following a more streamlined approach akin to the CRN method, balancing efficiency and effectiveness in various optimization settings.

Quasi-Newton methods, compared to Newton's method, approximate the Hessian matrix and its inverse instead of direct computation, which employ low-rank updates to significantly reduce the cost per iteration. Notable quasi-Newton formulas for minimization include the Broyden-Fletcher-Goldfarb-Shanno (BFGS)~\citep{broyden1970convergence, fletcher1970new, goldfarb1970family, shanno1970conditioning}, Davidon-Fletcher-Powell (DFP)~\citep{fletcher1970new, fletcher1963rapidly}, and Powell-Symmetric-Broyden (PSB)~\citep{powell1970new}. Despite their success in minimization problems, quasi-Newton methods are less commonly applied to minimax problems. Other advancements in quasi-Newton methods have been notable. For instance, \cite{rodomanov2021greedy, rodomanov2021new, rodomanov2022rates} introduced greedy and random variants that achieve non-asymptotic superlinear convergence. Following this, \cite{lin2022explicit} reported improvements in convergence rates that are independent of condition numbers, a significant enhancement over traditional models. Furthermore, research by \cite{jin2023non} has confirmed that non-asymptotic superlinear convergence also applies to the classical DFP, BFGS, and SR1 methods, solidifying their utility in complex optimization scenarios.

Despite their success in minimization problems, quasi-Newton methods are less commonly applied to minimax problems. Some research has explored proximal quasi-Newton methods for monotone variational inequalities ~\citep{burke2000superlinear, burke1999variable, chen1999proximal}, targeting convex-concave minimax problems. However, these adaptations may lack stability, particularly in simple bilinear settings, as indicated by numerical experiments. Recent developments have introduced new quasi-Newton methods tailored for minimax problems~\citep{abdi2019globally, essid2023implicit}, but comprehensive convergence rates for these methods have not been well documented.

In this paper, we explore quasi-Newton methods tailored for strongly-convex-strongly-concave saddle point problems. Specifically, we propose a multiple greedy quasi-Newton algorithm, which takes leverage of approximating the squared Hessian matrix with multiple greedy quasi-Newton updates per iteration. We rigorously establish a linear to quadratic convergence rate of our algorithm. Through numerical experiments on popular machine learning problems including AUC maximization and adversarial debiasing, we demonstrate the superior performance of our algorithm compared to state-of-the-art algorithms. The paper is organised as follows:

$\mathbf{Paper\ Organization}$ In Section~\ref{sec: notation}, we clarify the notations and provide assumptions and preliminaries of this paper. In Section~\ref{sec: method}, we introduce a framework for saddle point problems and propose our MGSR1-SP algorithm with theoretical convergence guarantee. In Section~\ref{sec: numerical}, we validate our algorithm on AUC maximization and Adverasial Debiasing tasks.

\section{Notation and Preliminaries}
\label{sec: notation}
We use $\|\cdot\|$ to denote the spectral norm and the Euclidean norm of a matrix and a vector, respectively. The standard basis in $\bbR^d$ is denoted by $\{\mathbf{e}_1, \dots, \mathbf{e}_d\}$. The identity matrix is represented as $\I$, and the trace of any square matrix is represented by $\text{tr}(\cdot)$. We use $\bbS_{++}^d$ to represent the set of positive definite matrices. For two positive definite matrices $\Q \in \bbS_{++}^d$ and $\H \in \bbS_{++}^d$, their inner product is defined as $\langle \Q, \H \rangle = \text{tr}(\Q\H)$. We denote $\Q \succeq \H$ if $\G - \H \succeq 0$. Referring to the objective function in equation~\eqref{eq: objective function}, let $d = d_\x + d_\y$ represent the full dimension. The gradient $\g(\x_k, \y_k)$ and Hessian matrix $\tilde{\H}(\x_k, \y_k)$ of function $f$ at the $k$-th iteration at $(\x_k, \y_k)$ are denoted as $\g_k \in \mathbb{R}^d$ and $\tilde{\H}_k \in \mathbb{R}^{d \times d}$ respectively, abbreviated for convenience. We also use $\H_{\x\x}, \H_{\x\y}$ and $\H_{\y\y}$ to denote the sub-matrices.\\\\
Suppose that the objective function in Eq.~\eqref{eq: objective function} satisfies the following assumptions:
\begin{assumption}
\label{assumption: lipschitz continuous}
The objective function $f(\mathbf{x}, \mathbf{y})$ is twice differentiable with an $L_1$-Lipschitz gradient and an $L_2$-Lipschitz Hessian, i.e., 
\begin{align*}
\|\mathbf{g}(\x, \y) - \g(\x', \y')\| \leq L_1\bigg\|\begin{bmatrix}\x - \x' \\ \y - \y'\end{bmatrix}\bigg\|
\end{align*}
and 
\begin{align*}
\|\tilde{\H}(\x, \y) - \tilde{\H}(\x', \y')\| \leq L_2\bigg\|\begin{bmatrix}\x - \x' \\ \y - \y'\end{bmatrix}\bigg\|
\end{align*}
for any $[\x; \y]^\top \in \bbR^{d} \text{ and } [\x'; \y']^\top \in \bbR^{d}$.
\end{assumption}

\begin{assumption}
\label{assumption: SCSC}
The objective function $f(\mathbf{x}, \mathbf{y})$ is $\mu$-strongly-convex in $\mathbf{x}$ and $\mu$-strongly-concave in $\mathbf{y}$, i.e., 
\begin{align*}
\tilde{\H}_{\x\x} \succeq \mu \I
\end{align*}
and 
\begin{align*}
\tilde{\H}_{\y\y} \preceq -\mu \I
\end{align*}
for any $[\x; \y]^\top \in \bbR^{d}$. Additionally, the condition number of the objective function is defined as $\kappa = L_1 / \mu$.
\end{assumption}

Note that the Hessian matrix $\tilde{\mathbf{H}}(\mathbf{x}, \mathbf{y})$ in saddle point problems in Eq.~\eqref{eq: objective function} is usually indefinite. However, the following lemma derived a crucial property of the squared Hessian matrix, providing a guarantee of positive definiteness.
\begin{lemma}
\label{lemma: squared Hessian}[\cite{liu2021quasi}]
Define $\H(\x, \y) = \tilde{\H}(\x, \y)^2$. Under Assumption~\ref{assumption: lipschitz continuous} and Assumption~\ref{assumption: SCSC}, we have $\mu^2 \mathbf{I} \preceq \H(\x, \y) \preceq L_1^2 \I$ for any $[\x; \y]^\top \in \bbR^d$.
\end{lemma}

\section{Methodology and Theoretical Analysis}
\label{sec: method}
In this section, we explore an innovative framework designed to address saddle point problems as described in Eq.~\eqref{eq: objective function}. We then review the basic principles of quasi-Newton methods, with a focus on the greedy variant discussed in \cite{rodomanov2021greedy}. Following this, we introduce our MGSR1-SP algorithm, which is characterized by its established linear-quadratic convergence rate and its near independence from condition numbers.

\subsection{A Quasi-Newton Framework for Saddle Point Problems}\label{subsec: Saddle Point Framework}
The standard update formula for Newton’s method is expressed as
\begin{align*}
\begin{bmatrix}
\x_{k+1} \\
\y_{k+1}
\end{bmatrix} = 
\begin{bmatrix}
\x_k \\
\y_k
\end{bmatrix} - \tilde{\H}_k^{-1} \g_k,
\end{align*}
which exhibits quadratic local convergence. However, this method incurs a computational complexity of $\calO(d^3)$ per iteration for the inverse Hessian matrix. In the realm of convex minimization, quasi-Newton methods such as BFGS, SR1, and their variations focus on approximating the Hessian matrix to reduce computational demands to $\calO(d^2)$ per iteration. Nonetheless, these methods presuppose a positive definite Hessian, which is unsuitable for saddle point problems as described in Eq.~\eqref{eq: objective function} due to the inherent indefiniteness of $\tilde{\H}(\x, \y)$. To address this challenge, \cite{liu2022quasi} reformulated the Newton's method as
\begin{align}
\label{eq:quasi1}
\begin{bmatrix}
\x_{k+1} \\
\y_{k+1}
\end{bmatrix} &= 
\begin{bmatrix}
\x_{k} \\
\y_{k}
\end{bmatrix} - [(\tilde{\H}_k)^2]^{-1} \tilde{\H}_k \g_k \notag\\
&= 
\begin{bmatrix}
\x_{k} \\
\y_{k}
\end{bmatrix}- \H_k^{-1} \tilde{\H}_k \g_k.
\end{align}
where $\H_k = \tilde{\H}_k^2$ is the auxiliary matrix defined in Lemma~\ref{lemma: squared Hessian}, which is guaranteed to be positive definite. Consequently, the update rule for Newton’s method can be reformulated as
\begin{align}\label{eq:quasi2}
\begin{bmatrix}
\x_{k+1} \\
\y_{k+1}
\end{bmatrix} &= \begin{bmatrix}
\x_{k} \\
\y_{k}
\end{bmatrix} - \Q_k^{-1} \tilde{\H}_k\g_k,
\end{align}
where $\Q_k^{-1} \in \bbS_{++}^d$ is an approximated inverse matrix of $\H_k^{-1}$. We will introduce the techniques to construct $\Q_k$ and its inverse in the next section. Note that the update rule~\eqref{eq:quasi2} does not necessarily require the explicit construction of Hessian matrix, which can be computed efficiently through Hessian-Vector Product (HVP)~\citep{pearlmutter1994fast, schraudolph2002fast}.

\subsection{Greedy Quasi-Newton Methods}
\label{subsec: Quasi-Newton Basics}
Quasi-Newton methods are developed to circumvent some of the computational inefficiencies associated with the classical Newton's method~\citep{broyden1967quasi, shanno1970conditioning, broyden1970convergence, broyden1973local, davidon1991variable, lee2018distributed, rodomanov2021greedy, liu2022quasi, ye2021explicit, lin2022explicit,rodomanov2022rates, jin2023non}. Among these, the Broyden family updates, particularly the Broyden-Fletcher-Goldfarb-Shanno (BFGS) formula, has become the most celebrated, widely cited in many literature. Given two symmetric positive definite matrices $\H, \Q \in \bbS^{d}_{++}$ and a vector $\u \in \bbR^{d}$, satisfying that $\Q \succeq \H$ and $\Q\u \neq \A\u$, the Broyden family update is given by
\begin{align*}
\text{Broyd}_\tau(\Q, \H, \u) = \tau \text{DFP}(\Q,&\H, \u) \\ &+ (1 - \tau) \text{SR1}(\Q, \H, \u),
\end{align*}
which is proved to be a convex combination~\citep{rodomanov2021greedy}. The SR1 update refines the Hessian approximation by
\begin{align*}
\text{SR1}(\Q, \H, \u) = \Q - \frac{(\Q - \H)\u\u^\top(\Q-\H)}{\u^\top (\Q - \H)\u},
\end{align*}
which leverages a rank-one modification to adjust $\Q$ based on the discrepancy between $\Q$ and $\H$. On the other hand, the DFP update incorporates both current and previous curvature information into the approximation:
\begin{align*}
\text{DFP}(\Q, \H, \u) = \Q - &\frac{\H\u\u^\top \Q + \Q\u\u^\top \H}{\u^\top \H\u} \\
&+ \bigg(1 + \frac{\u^\top \Q\u}{\u^\top \H\u}\bigg)\frac{\H\u\u^\top\H}{\u^\top \H\u}.
\end{align*}
The parameter $\tau \in [0, 1]$ determines the specific type of Quasi-Newton update applied. Specifically, setting $\tau = \frac{\u^\top \H\u}{\u^\top \Q\u}$ leads to the well-known Broyden-Fletcher-Goldfarb-Shanno (BFGS) update formulated as
\begin{align*}
\text{BFGS}(\Q, \H, \u) = \Q - \frac{\Q\u\u^\top \Q}{\u^\top \Q\u} + \frac{\H\u\u^\top \H}{\u^\top \H\u}.
\end{align*}

Greedy Quasi-Newton methods were proposed to achieve better convergence rates compared to classical Quasi-Newton methods, with a contraction factor that depends on the square of the iteration counter \citep{rodomanov2021greedy}. Specifically, for a given target matrix $\H \in \bbS_{++}^d$ and an approximator $\Q \succeq \H$, the greedily selected vector $\u$ is determined as follows:
\begin{align*}
\u_\H(\Q) = \argmax_{\u \in \{\e_1, \ldots, \e_n\}} \frac{\u^\top \Q\u}{\u^\top \H\u},
\end{align*}
where $\e_i$ represents the basis vector. Define the greedy Broyden family update as follows:
\begin{align*}
\text{gBroyd}_\tau(\Q, \H) \defeq \text{Broyd}_\tau(\Q, \H, \u_\H(\Q)).
\end{align*}
Specifically, if $\tau = 0$, the update is greedy SR1 update defined as
\begin{align}
\label{eq: gSR1}
\text{gSR1}(\Q, \H) \defeq \text{gSR1}(\Q, \H, \u_\H(\Q)).
\end{align}
The following lemma demonstrates that the greedy SR1 update reduces the rank of $\Q - \H$ at each iteration. Therefore, with at most $d$ iterations, the gSR1 update will accurately recover the Hessian matrix.
\begin{theorem}[\cite{rodomanov2021greedy}, Theorem 3.5]
Suppose that for each $k \geq 0$, we choose $\u_k = \u_\H(\Q_{k})$ and $\tau = 0$, then $\Q_k = \H$ for some $0 \leq k \leq d$.
\end{theorem}
In quadratic optimization, the Hessian matrix remains constant, and with each iteration, the approximated Hessian matrix converges towards the true Hessian, as demonstrated by the previous lemma. For more general problems, we define the multiple $\text{gBroyd}_\tau^n$ as a series of nested $\text{gBroyd}\tau$ updates~\citep{du2024distributed}, targeting the same Hessian matrix $\H$, such that
\begin{align*}
\Q_{i+1} \leftarrow \text{gBroyd}_\tau^n(\Q_i, \H), \quad i = 0, \dots, n-1,
\end{align*}
where $n$ is a non-negative integer representing the number of rounds of greedy Broyden family updates performed in each iteration. Specifically, for the multiple greedy SR1 updates denoted as $\text{gSR1}^n$, within each iteration, the updates occur as follows:
\begin{align}
\label{eq: mgSR1}
\Q_{i+1} \leftarrow \text{gSR1}(\Q, \H), \quad i = 0, \dots, n-1,
\end{align}

\subsection{MGSR1-SP Algorithm and Convergence Analysis}
\label{subsec: our algorithm}
In this section, we introduce the Multiple Greedy Rank-1 (MGSR1-SP) algorithm for solving Saddle Point Problems satisfying Assumption~\ref{assumption: lipschitz continuous} and Assumption~\ref{assumption: SCSC}, which is outlined in Algorithm~\ref{algo: MGSR1-SP} with established convergence guarantee. The MGSR1-SP algorithm builds upon the framework in Section~\ref{subsec: Saddle Point Framework} and adopts the multiple greedy SR1 updates specified in Eq.~\eqref{eq: mgSR1}.
\begin{algorithm}
\caption{MGSR1-SP}
\label{algo: MGSR1-SP}
\begin{algorithmic}[1]
\State \textbf{Initialization:} $\z_0$, $\Q_0$, stepsize $\alpha$, $M$, and $n \geq 0$.
\For{k \text{in} $0, \dots, N$}
    \State Compute $\g_k$\vskip 0.1cm
    \State Update $\z_{k+1} \leftarrow \z_k - \alpha \cdot \Q_k^{-1} \tilde{\H}_k \g_k\ \text{(HVP)}$ \vskip 0.1cm
    \State Perform $\text{gSR1}^n$ updates:
    $\tilde{\Q}_k = \text{gSR1}^n(\Q_k, \H_k).$ \vskip 0.1cm
    \State Compute $r_{k} = \bigg\|\begin{bmatrix} \x_{k+1} - \x_k \\ \y_{k+1} - \y_k\end{bmatrix}\bigg\|$ \vskip 0.1cm
    \State Correct $\tilde{\Q}_{k+1} \leftarrow (1 + Mr_k)\tilde{\Q}_k$ \vskip 0.1cm
    \State Compute $\Q_{k+1} = \text{gSR1}(\tilde{\Q}_{k+1}, \H_{k+1}))$ \vskip 0.1cm
\EndFor
\end{algorithmic}
\end{algorithm}

\begin{lemma}[Modified from \cite{rodomanov2021greedy}]
\label{lemma: update lemma}
If, for some $\eta \geq 1$, and two positive definite matrix $\H, \Q \in \bbS_{++}^d$, we have
\begin{align*}
\H \preceq \Q \preceq \eta \H,
\end{align*}
then using greedy SR1 update~\eqref{eq: gSR1}, we also have
\begin{align*}
\H \preceq \text{gSR1$(\Q, \H)$} \preceq \eta \H.
\end{align*}
\end{lemma}

\begin{lemma}[Modified from \citep{liu2021quasi}]
\label{lemma: squared matrix iteration}
Let $[\x_k; \y_k]^\top, [\x_{k+1}; \y_{k+1}]^\top \in \bbR^{d}$ with squared Hessian matrix $\H_k, \H_{k+1} \in \bbS_{++}^d$ defined in Lemma~\ref{lemma: squared Hessian}. For some $\eta \geq 1 $ and let $\Q_k \in \bbS_{++}^d$ be a positive definite matrix such that 
\begin{align*}
\H_k \preceq \Q_k \preceq \eta \H_k,
\end{align*}
we have 
\begin{align*}
\H_{k+1} \preceq \text{gSR1}(\tilde{\Q}_{k+1}, \H_{k+1}) \preceq (1+Mr_k)^2 \eta \H_{k+1} 
\end{align*}
where $\tilde{\Q}_{k+1} = (1+Mr_k)\Q_k$, $r_k = \bigg\|\begin{bmatrix} \x_{k+1} - \x_k \\ \y_{k+1} - \y_k\end{bmatrix}\bigg\|$ and  $M = \frac{2\kappa^2 L_2}{L_1}$.
\end{lemma}

Given upon this, define the convergence measure as 
\begin{align}
\label{eq: convergence measure}
\lambda_k = \lambda(\x_k, \y_k) = \|\g(\x_k, \y_k)\|_2,
\end{align}
we establish a linear to quadratic convergence rate for our MGSR1-SP algorithm in the following theorem:
\begin{theorem}
\label{thm: convergence}
Using Algorithm~\ref{algo: MGSR1-SP}, suppose we have $\H_k \preceq \Q_k \preceq \eta_k \H_k$ for some $ \eta_k \geq 1 $, and let $ \beta = \frac{L_2}{2\mu^2} $, then we have
\begin{align*}
\lambda_{k+1} &\leq \bigg(1 - \frac{1}{\eta_k}\bigg) \lambda_k + \beta \lambda_k^2.
\end{align*}
\end{theorem}
\begin{proofsketch}
Suppose $\H_k \preceq \Q_k \preceq \eta_k\H_k$ holds, with Lemma~\ref{lemma: update lemma}, the multiple greedy SR1 update also satisfies $\H_k \preceq \text{gSR1}^n(\Q_k, \H_k) \preceq \eta \H_k$, hence, following Lemma~\ref{lemma: squared matrix iteration}, we have $\H_{k+1} \preceq \Q_{k+1} \preceq (1 + Mr_k)^2\eta_k \H_{k+1} = \eta_{k+1}\H_{k+1}$. The rest follows Lemma 3.14 in \cite{liu2021quasi}.
\end{proofsketch}

\section{Numerical Experiments}
\label{sec: numerical}
In this section, we demonstrate the efficiency of our algorithm using two popular machine learning tasks: AUC maximization and adversarial debiasing. The experiments are conducted on a Macbook Air with M2 chip.

\subsection{AUC Maximization}
In machine learning, the Area Under the ROC Curve (AUC) is a key metric that evaluates classifier performance in binary classification, particularly useful with imbalanced data. The problem can be formulated as follows:
\begin{align*}
f(\x, \y) := \frac{1}{m} \sum_{i=1}^{m} f_i(\x, y) + \frac{\lambda}{2} \|\x\|^2 - p(1-p)y^2,
\end{align*} 
where $\x = [\w; u; v]^\top$, $\lambda$ is the regularization parameter and $p$ denotes the proportion of positive instances in the dataset. The function $f_i(\x, y)$ is defined as:
\begin{align*}
f_i(\x,y) = &(1-p)\big((\w^\top \a_j-u)^2 - 2(1+y)\w^\top \a_j\big)\I_{b_j=1}\\
&+p\big((\w^\top \a_j-v)^2 + 2(1+y)\w^\top \a_j\big)\I_{b_j=-1}, 
\end{align*}
where \(\a_i \in \mathbb{R}^{n_\x-2}\) are features and \(b_i \in \{+1, -1\}\) is the label.

\begin{figure}\centering
\begin{tabular}{cc}
\includegraphics[scale=0.24]{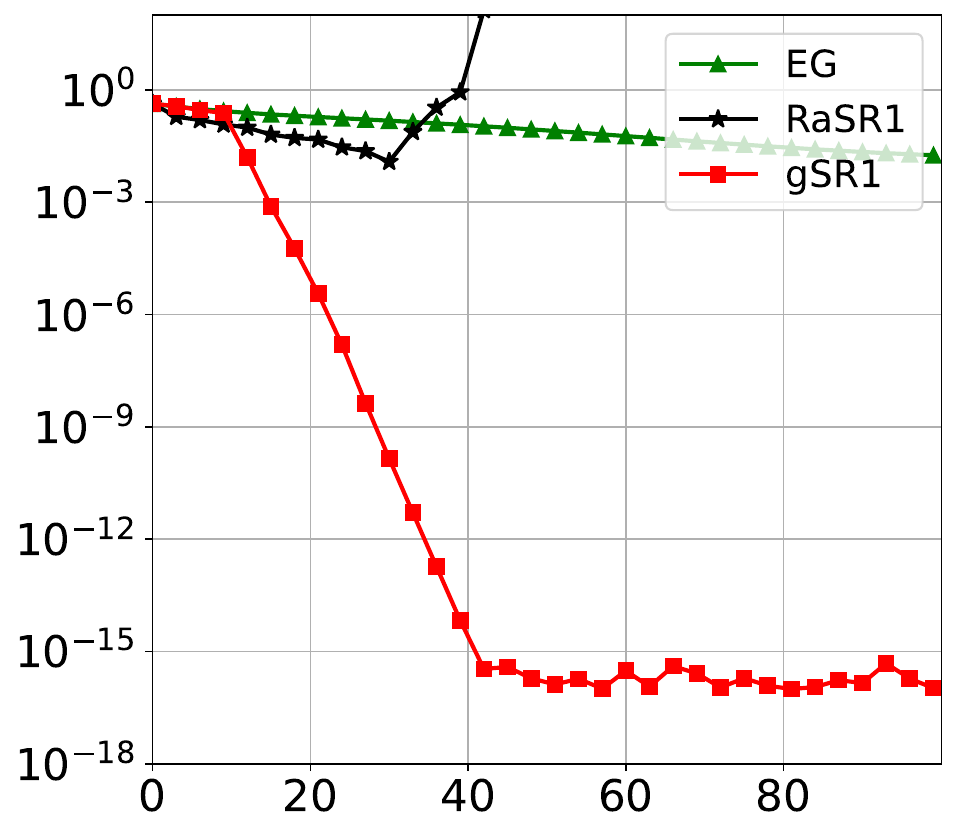} & 
\includegraphics[scale=0.24]{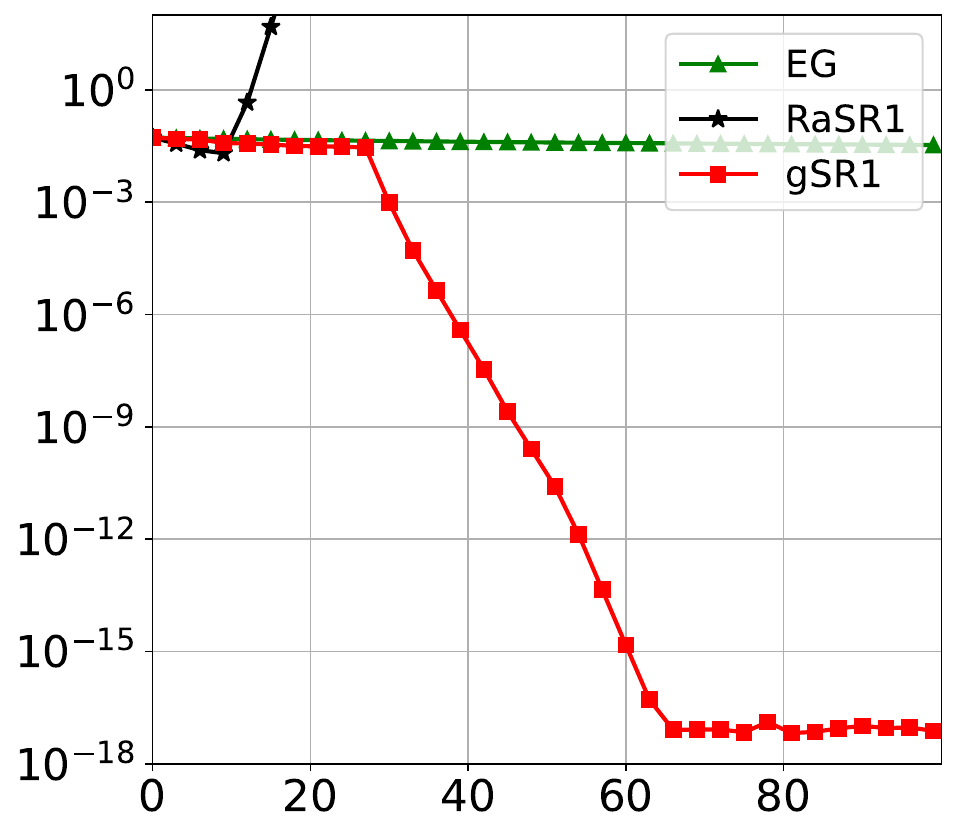}\\[-0.1cm]
(a) a9a, \#iter & (b) w8a, \#iter\\[0.1cm]
\includegraphics[scale=0.24]{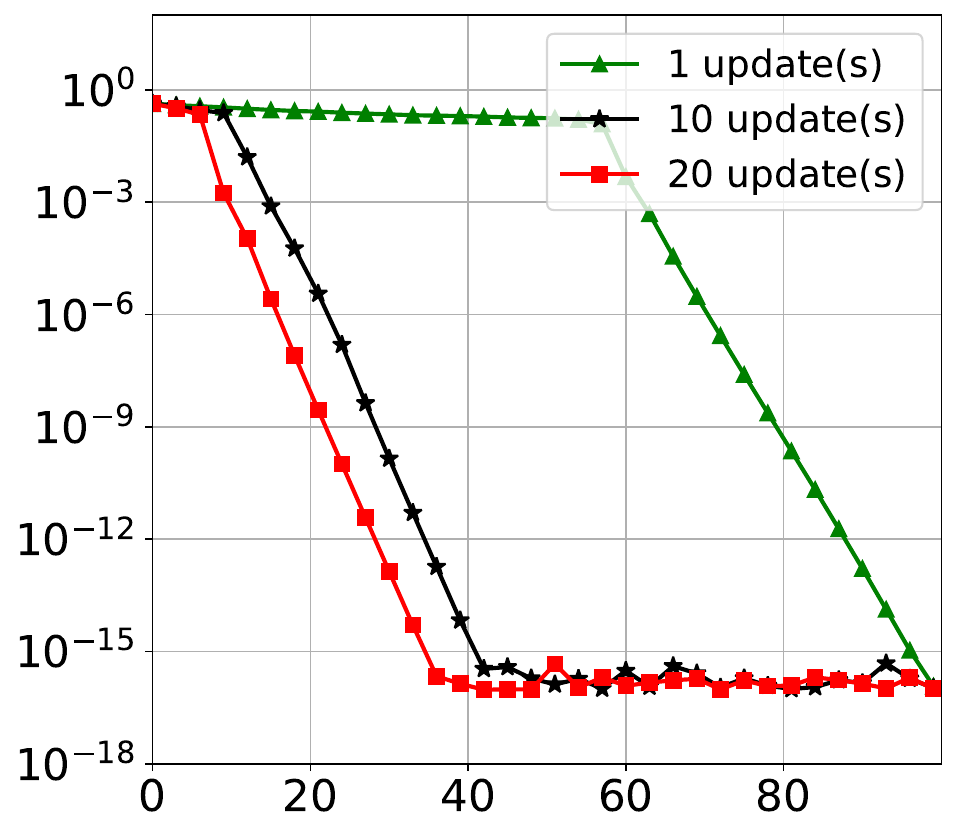} &
\includegraphics[scale=0.24]{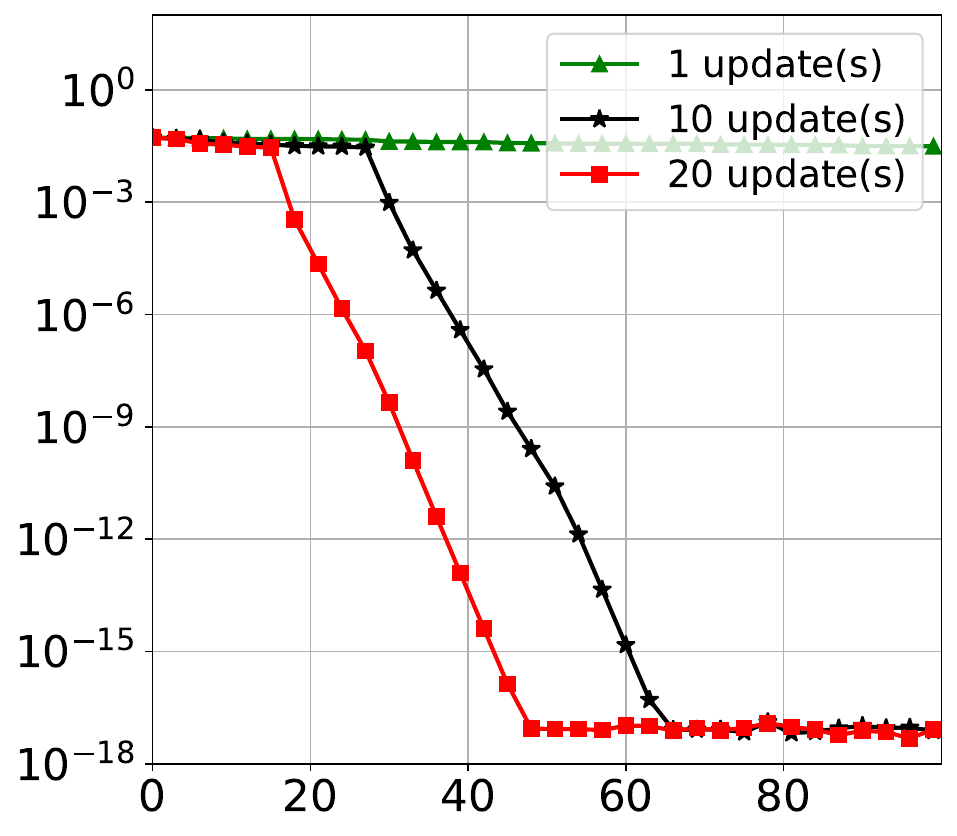} \\[-0.1cm]
(c) a9a, \#iter & (d) w8a, \#iter
\end{tabular}
\caption{Numerical results for AUC Maximization task. The y-axis denotes the gradient norm $\|\nabla f(\x,y)\|_2$ and x-axis denotes the number of iterations. Top two figures compares Extragradient, RandomSR1 and MGSR1-SP(`gSR1') with 20 rounds update on `a9a' and `w8a' dataset. Bottom two figures compare MGSR1-SP(`gSR1') with different number of updates per iteration.}\vskip-0.3cm
\label{fig: auc}
\end{figure}

\subsection{Adversarial Debiasing}
Adversarial debiasing is a prominent method used to enhance equity in AI by integrating adversarial techniques to mitigate biases within machine learning algorithms. Given a dataset $\{\a_i, b_i, c_i\}_{i=1}^{m}$, where $\a_i$ represents input variables, $b_i \in \mathbb{R}$ is the output, and $c_i \in \mathbb{R}$ is the protected variable, the objective is to reduce bias, which can be formulated as:
\begin{align*}
f(\x, y) = \frac{1}{m} \sum_{i=1}^{m} f_i(\x, y) + \lambda \|\x\|^2 - \gamma y^2 ,
\end{align*} 
where $\lambda, \gamma$ are regularization parameters. Function $f_i(\x, y)$ is defined as
\begin{align*}
f_i(\x, y) = &\log\big(1+\exp\big(-b_j(\a_j)^\top\x\big)\big) \\
&-\beta\log\big(1 + \exp\big(-c_j(\a_j)^\top\x y\big)\big), 
\end{align*}
with $\beta$ also serving as a regularization parameter.

\begin{figure}\centering
\begin{tabular}{cccc}
\includegraphics[scale=0.24]{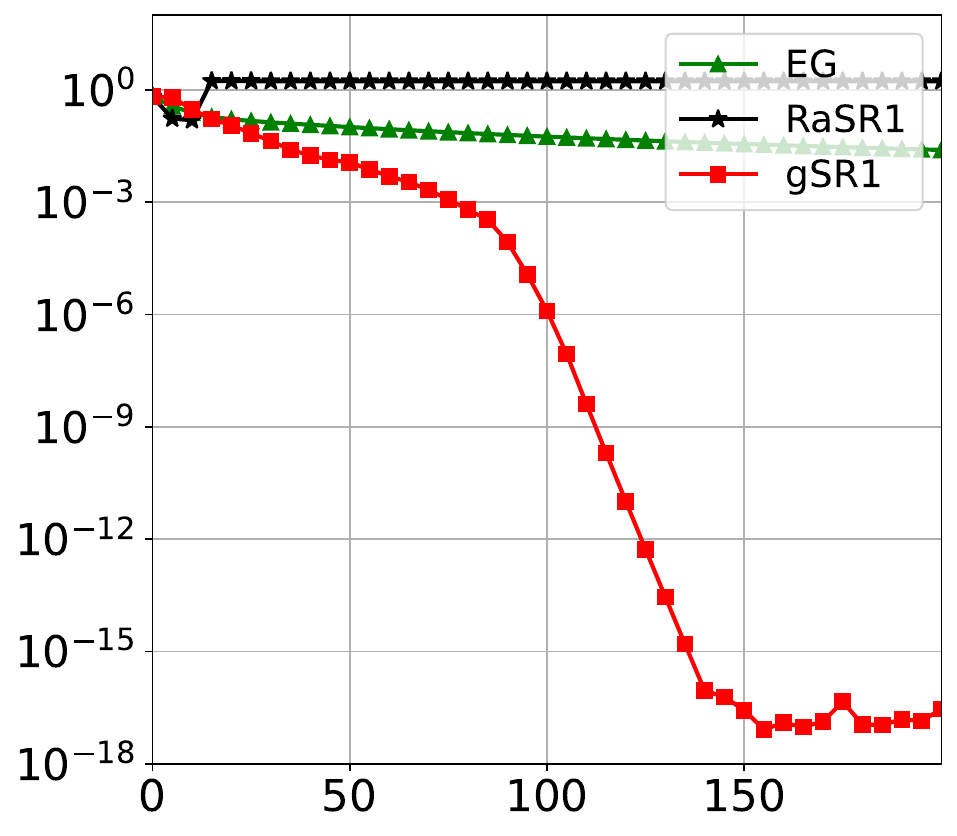} & 
\includegraphics[scale=0.24]{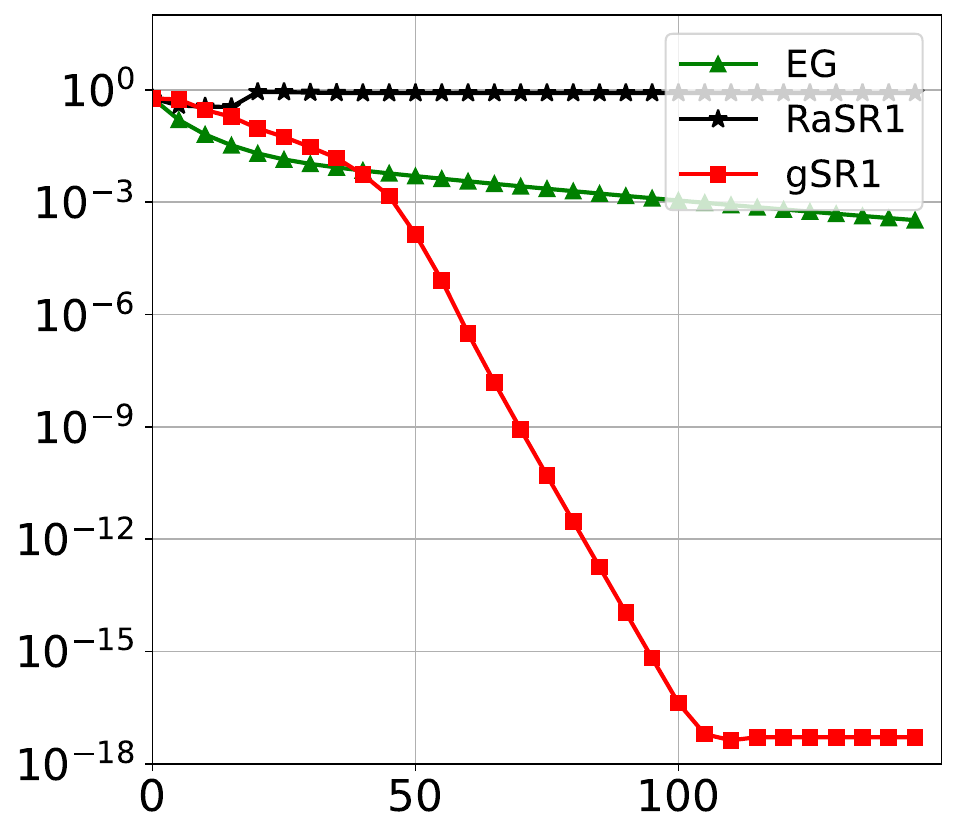} \\[-0.1cm]
(a) adult, \#iter & (b) law, \#iter \\[0.1cm]
\includegraphics[scale=0.24]{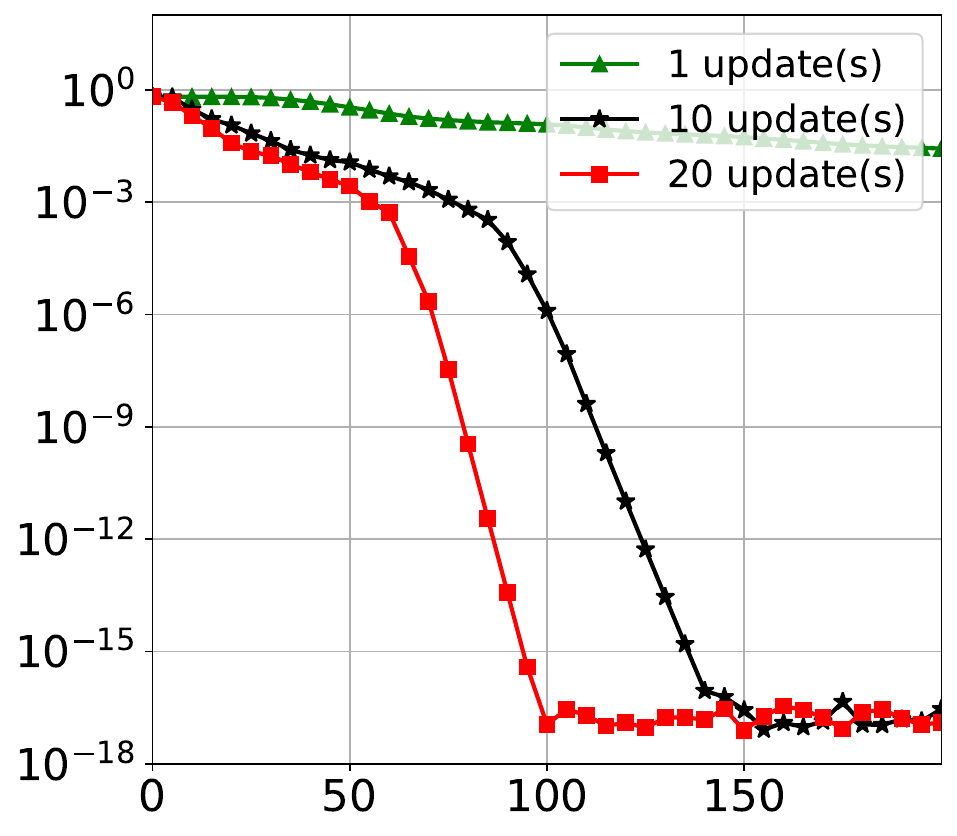} & 
\includegraphics[scale=0.24]{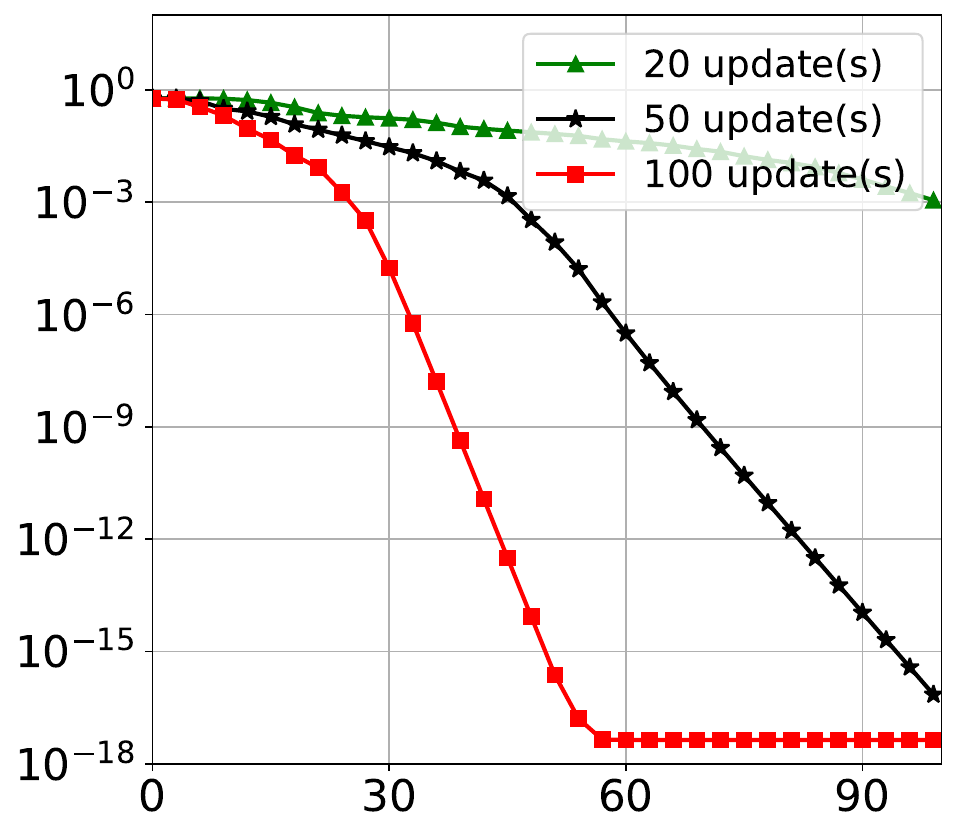} \\[-0.1cm]
(c) adult, \#iter & (d) law, \#iter
\end{tabular}
\caption{Numerical results for Adversarial Debiasing. The y-axis denotes the gradient norm $\|\nabla f(\x,y)\|_2$ and x-axis denotes the number of iterations. Top two figures compares Extragradient, RandomSR1 and MGSR1-SP(`gSR1') with 20 rounds update on `a9a' and `w8a' dataset. Bottom two figures compare MGSR1-SP(`gSR1') with different number of updates per iteration.}\vskip-0.3cm
\label{fig: fairness}
\end{figure}

\subsection{Analysis}
We evaluated the performance of our MGSR1-SP algorithm against two baselines: Random SR1, where vectors $\mathbf{u} \in \mathbb{R}^d$ are drawn from a normal distribution $\mathcal{N}(0, 1)$, and the ExtraGradient algorithm for saddle point problems.

For AUC maximization, the experiments were conducted on the `a9a' dataset ($n_{\mathbf{x}}=125, n_{\mathbf{y}}=1, N=32651$) and the `w8a' dataset ($n_{\mathbf{x}}=302, n_{\mathbf{y}}=1, N=45546$). The results are shown in Figure~\ref{fig: auc}. Notably, the Hessian AUC maximization is invariant ($L_2 = 0$), indicating a linear convergence rate~\ref{thm: convergence}. Our MGSR1-SP algorithm demonstrated a faster convergence rate compared to the ExtraGradient algorithm. Moreover, it offered more stable Hessian approximations than the random SR1 update, particularly as the number of update rounds increased. 

For adversarial debiasing, the experiments were conducted using the `adult' dataset ($n_{\mathbf{x}}=122, n_{\mathbf{y}}=1, N=32651$) and the `law school' dataset ($n_{\mathbf{x}}=379, n_{\mathbf{y}}=1, N=20427$). The results, shown in Figure~\ref{fig: fairness}, indicated that our algorithm achieved a linear-quadratic convergence rate, supporting Theorem~\ref{thm: convergence}. Our method outperformed both ExtraGradient and Random SR1 in terms of iterations required, with significant performance improvements as updates increased.

\section{Conclusion}
In this paper, we introduce the Multiple Greedy Quasi-Newton (MGSR1-SP) method, a novel approach designed to solve strongly-convex-strongly-concave (SCSC) saddle point problems. This algorithm approximates the square of the indefinite Hessian matrix, enhancing accuracy and efficiency through a series of iterative, enhanced quasi-Newton updates. Our comprehensive convergence analysis rigorously establishes the theoretical results of the MGSR1-SP algorithm, demonstrating its linear-quadratic convergence rates. Furthermore, we conducted extensive empirical validations against state-of-the-art optimization methods, including the ExtraGradient and Random SR1 algorithms, on two popular machine learning applications: AUC maximization and adversarial debiasing. The results clearly show that our method not only converges faster but also provides a more accurate estimation of the Hessian inverse, leading to more stable and reliable optimization outcomes.

For future work, several promising directions can be explored. These include adapting the MGSR1-SP framework to stochastic settings. Additionally, the development of limited memory quasi-Newton methods could make our approach feasible for large-scale problems, where computational resources and memory usage are significant constraints. Another area of potential exploration is the integration of adaptive step-size selection mechanisms to enhance effectiveness. Lastly, extending our method to tackle non-convex saddle point problems with appropriate regularization could broaden its applicability to a wider array of problems in machine learning and beyond.

\section{Acknowledgment}
The first paragraph of the conclusion section in this manuscript was polished using ChatGPT to enhance clarity and readability. The authors take full responsibility for the content of this manuscript, including this AI-assisted revision. 
\bibliographystyle{unsrtnat}
\bibliography{references} 
\end{document}